\documentclass[lettersize,journal, onecolumn]{IEEEtran}
\usepackage{amsmath,amssymb,amsfonts}
\usepackage{array}
\usepackage{textcomp}
\usepackage{stfloats}
\usepackage{url}
\usepackage{verbatim}
\usepackage{graphicx}
\usepackage{cite}
\usepackage{amssymb}
\usepackage{mathtools}
\usepackage{multicol}
\usepackage{multirow,tabularx}
\usepackage{booktabs}
\usepackage{caption}
\usepackage{subfigure}
\usepackage{hyperref}
\usepackage{algorithm2e}

\usepackage{hyperxmp}
\usepackage{graphicx}
\usepackage{hyperref}
\usepackage{array}
\usepackage{amsmath}

\usepackage{pifont}
\usepackage{tikz}
\usetikzlibrary{positioning}
\usepackage{amssymb}
\usepackage{multirow,tabularx}
\usepackage{multicol}
\usepackage{booktabs}
\usepackage{enumitem}
\usepackage[para,online,flushleft]{threeparttable}

\begin{document}

\title{\textsc{HalluShift++}: Bridging Language and Vision through Internal Representation Shifts for Hierarchical Hallucinations in MLLMs}

\author{Sujoy Nath, Arkaprabha Basu, Sharanya Dasgupta, and Swagatam Das
\IEEEcompsocitemizethanks{\IEEEcompsocthanksitem Sujoy Nath (sujoynathofficial@gmail.com), Arkaprabha Basu, Sharanya Dasgupta, and Swagatam Das (swagatam.das@isical.ac.in) are with Netaji Subhash Engineering College (NSEC), Kolkata, India, the TCG Crest, Kolkata, India and the Electronics and Communication Sciences Unit (ECSU), Indian Statistical Institute, Kolkata, India.  \protect
\IEEEcompsocthanksitem Corresponding author: Swagatam Das.}}

\maketitle

\begin{abstract}
Multimodal Large Language Models (MLLMs) have demonstrated remarkable capabilities in vision-language understanding tasks. While these models often produce linguistically coherent output, they often suffer from hallucinations, generating descriptions that are factually inconsistent with the visual content, potentially leading to adverse consequences. Therefore, the assessment of hallucinations in MLLM has become increasingly crucial in the model development process. Contemporary methodologies predominantly depend on external LLM evaluators, which are themselves susceptible to hallucinations and may present challenges in terms of domain adaptation. In this study, we propose the hypothesis that hallucination manifests as measurable irregularities within the internal layer dynamics of MLLMs, not merely due to distributional shifts but also in the context of layer-wise analysis of specific assumptions. By incorporating such modifications, \textsc{\textsc{HalluShift++}} broadens the efficacy of hallucination detection from text-based large language models (LLMs) to encompass multimodal scenarios. Our codebase is available at \url{https://github.com/C0mRD/HalluShift_Plus}.
\end{abstract}

\begin{IEEEkeywords}
Hallucination Detection, Multi-modal Large Language Models, Object Hallucination
\end{IEEEkeywords}
\begin{figure}[htb]
    \centering
    \includegraphics[width=0.7\columnwidth]{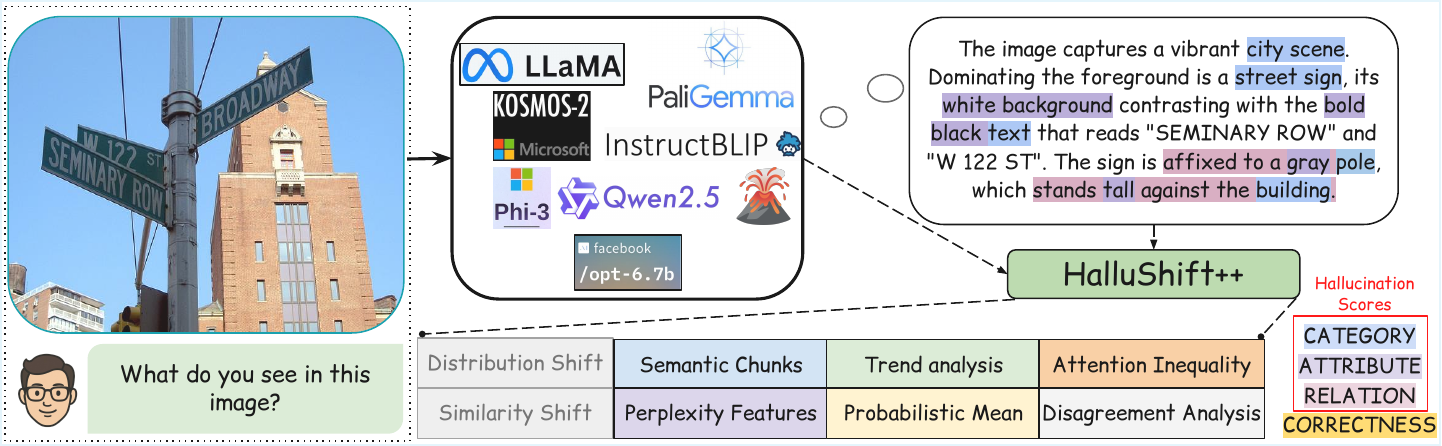} 
    \caption{\textsc{HalluShift++} in action: Left: MLLMs generate descriptions for visual inputs that may contain hallucinations. Right: Our framework provides hierarchical hallucination scores across three types (Category, Attribute, Relation).}
    \label{fig:teaserfigure}
\end{figure}

\section{Introduction}
Nowadays, MLLMs \cite{liu2023visualinstructiontuning, beyer2024paligemma, pan2024kosmos, dai2023instructblip} show a promising ability to harness the language generation capabilities of LLMs \cite{achiam2023gpt, touvron2023llama} along with visual understanding to excel in multimodal tasks, such as image captioning \cite{li2023blip} and visual question answering \cite{liu2023visualinstructiontuning, dai2023instructblip}. These models are frequently susceptible to producing plausible sounding descriptions, therefore, being grammatically correct, but such descriptions often comes with inconsistent visual content, a phenomenon commonly referred to as hallucination \cite{li2023evaluating, liu2024survey, rohrbach2018object}. \\
\noindent
We have observed hallucinations in MLLMs exhibit two distinct failure modes: objects that are present in images are missing from description, and describing objects that are not present in a certain image. While demonstrating object hallucination, addressing the first type is beyound the scope of this work and our primary focus resides on the second type concerning correctness being vital to human judges rather than specificity \cite{rohrbach2016moviedescription}. Object hallucination can be further categorized into three types: object category, object attribute (color, shape, or counting), and object relation (human-object interactions or relative positions) \cite{bai2024hallucination, wu2024evaluating}. Hallucination degrades the model performance and greatly harm the user experiences in real-world applications \cite{ji2023survey}. \\ 
\noindent
Current approaches in hallucination detection often uses an external LLM like GPT-4 \cite{achiam2023gpt} as a judge \cite{chen2024unified, liu2023mitigating}. These models typically produce definitive answers as either \textit{hallucinated} or \textit{truthful}, but unable to precisely convey the nature of the hallucination, as evidenced by their confidence score at the token level. In contemporary research, it is apparent that there exists a significant demand for an effective hallucination detection mechanism that can be effortlessly incorporated into current MLLMs. Such a mechanism should additionally provide hallucination-type annotations, as various types necessitate distinct considerations regarding the training datasets and architectural design. \\
\noindent
Our central challenge is to meaningfully extract and analyze the foundational representations embedded inside the internal probability distributions of MLLMs that directly correlate with specific types of hallucinations. First, can we derive more robust and interpretable features from these internal distributions to reliably distinguish between content that is factually grounded and content that is hallucinated? And second, if content is indeed identified as hallucinated, can these same features further enable us to precisely classify the specific type of hallucination present of its nature? To delineate this challenge in MLLMs, we propose \textsc{\textsc{HalluShift++}}, an enhanced hallucination detection framework. While \textsc{HalluShift} \cite{dasgupta2025hallushift} works solely based on internal distribution shifts within language models, \textsc{HalluShift++} adapted the distribution nature of both LLM and MLLM with additional internal representation supports. Building upon the foundation of \textsc{HalluShift} \cite{dasgupta2025hallushift}, our approach extends the distributional shift analysis to the multi-modal domain where hallucinations arise from misaligned visual-linguistic representations and inadequate visual grounding. We summarize our key contributions as follows:
\begin{itemize}
    \item \textit{Extended \textsc{HalluShift} from LLMs to MLLMs}, we introduce 15 novel features (e.g. layer consistency, attention concentration) that capture previously unexplored aspects of multimodal hallucination patterns.
    \item A \textit{semantic chunking strategy} that automatically decomposes generated descriptions into semantically meaningful units (object, attribute, and relation chunks), enriched with contextual features, using part-of-speech analysis, named entity recognition \cite{keraghel2024ner}, and dependency parsing to enable precise data labeling across different hallucination categories.
    \item  \textit{Multi-class hallucination taxonomy} beyond binary hallucination detection, our membership function assigns probabilistic scores across 3 token-level object hallucination classes: category, attribute, and relation hallucination.
    \item We perform extensive experiments across various foundational MLLMs, resulting in highest improvement of $\sim64.12\%$ in AUC-ROC compared to previous \textsc{HalluShift} in image-captioning tasks. Additionally, we achieve $\sim3\%$ improvement on LLMs QA tasks.
\end{itemize}

\section{Related Works}
\textbf{Hallucination Detection Methods in MLLMs} Early works on hallucination detection in MLLMs focused primarily on detecting nonexisting objects in generated image captions. CHAIR \cite{bordes2024chair} employs a rule-based method to determine the proportion of words generated that correspond to the ground truth annotations and the segmentations of objects, showing limitations only in the detection of object-level hallucinations. In contrast, \cite{li2023evaluating} shows that the instruction design and the length of the generated captions can affect the performance of CHAIR. Thus, the authors propose a new evaluation metric, called POPE \cite{li2023evaluating}, transforming hallucination detection into a binary classification. This approach involves simply prompting MLLMs with multiple \textit{Yes} or \textit{No} questions blandly with all available classes about probing objects. Later approaches use LLMs as a judge to detect hallucinations. GAVIE \cite{liu2023mitigating} employs GPT-4 \cite{achiam2023gpt} to evaluate the instruction-following performance and assign a single accuracy score to responses. FAITHScore \cite{jing2023faithscore} aims to evaluate free-form responses to open-ended questions using a chain of large models to break down responses into atomic facts and verify via visual entailment. Furthermore, HaELM \cite{wang2023evaluation}, trained a LLM based on LLaMA for hallucination detection on a set of data generated by simulating different hallucination scenarios using ChatGPT. Although, such approaches can detect hallucination on long captions, they are incapacitated of providing hallucination type annotation. HalLoc \cite{park2025halloc} addresses this challenge by employing a token-level detection mechanism to identify hallucinations of objects, attributes, relationships, and scenes in various tasks. However, most of this available methods relies solely on external LLMs as evaluators, which again makes verification prone to hallucination. The evaluator is a novel network that has been trained on either substantially larger or distinct datasets, which invariably raises questions regarding the implications of the hallucinated model when iterating through its own internal distributions.\\
\noindent
\textbf{Internal Representation Analysis} Recent approaches have explored the internal states of LLM to detect hallucinations without relying on external evaluators. These methods operate on the hypothesis that hallucinated content exhibits distinctive patterns in model's internal representations that can be captured and analyzed \cite{li2023inference}. Techniques such as covariance pattern analysis \cite{chen2024inside} and eigenvalue decomposition of hidden states \cite{sriramanan2024llm} revealed latent level inconsistencies correlated with hallucinations. Haloscope \cite{du2024haloscope}, experimented with geometric subspace analysis of selected LLM layers, but it does not successfully capture any probabilistic features or shift in internal dynamics of LLMs, as it concentrates solely on hallucinations from a geometric transformation’s perspective. Taking a step forward, \textsc{HalluShift} \cite{dasgupta2025hallushift} originally developed for text-only LLMs, represents a notable contribution by hypothesizing that hallucinations manifest as quantifiable distribution shifts in the model's internal layer dynamics, extracting features from hidden and attention state spaces across LLM decoder layers. While \textsc{HalluShift} significantly outperforms existing SOTA methods across multiple benchmark on text-based QA tasks, it was limited to uni-modal text generation scenarios. Our observations indicate that \textsc{HalluShift} faces significant challenges when extended to multimodal contexts and predominately fails to capture complex visual-linguistic interactions, where hallucinations can emerge from inadequate visual grounding and language bias. In the context of multimodal large language models (MLLMs), research focusing on hallucination detection through the exploration of the models' internal representations is comparatively scarce. Conversely, there is a wealth of studies employing external LLM evaluators. This disparity highlights a significant research opportunity to develop specialized approaches that leverage internal representations, thereby effectively addressing the distinct challenges associated with detecting hallucinations in MLLMs.
\begin{figure*}[htbp]
  \centering
  \includegraphics[width=0.9\textwidth]{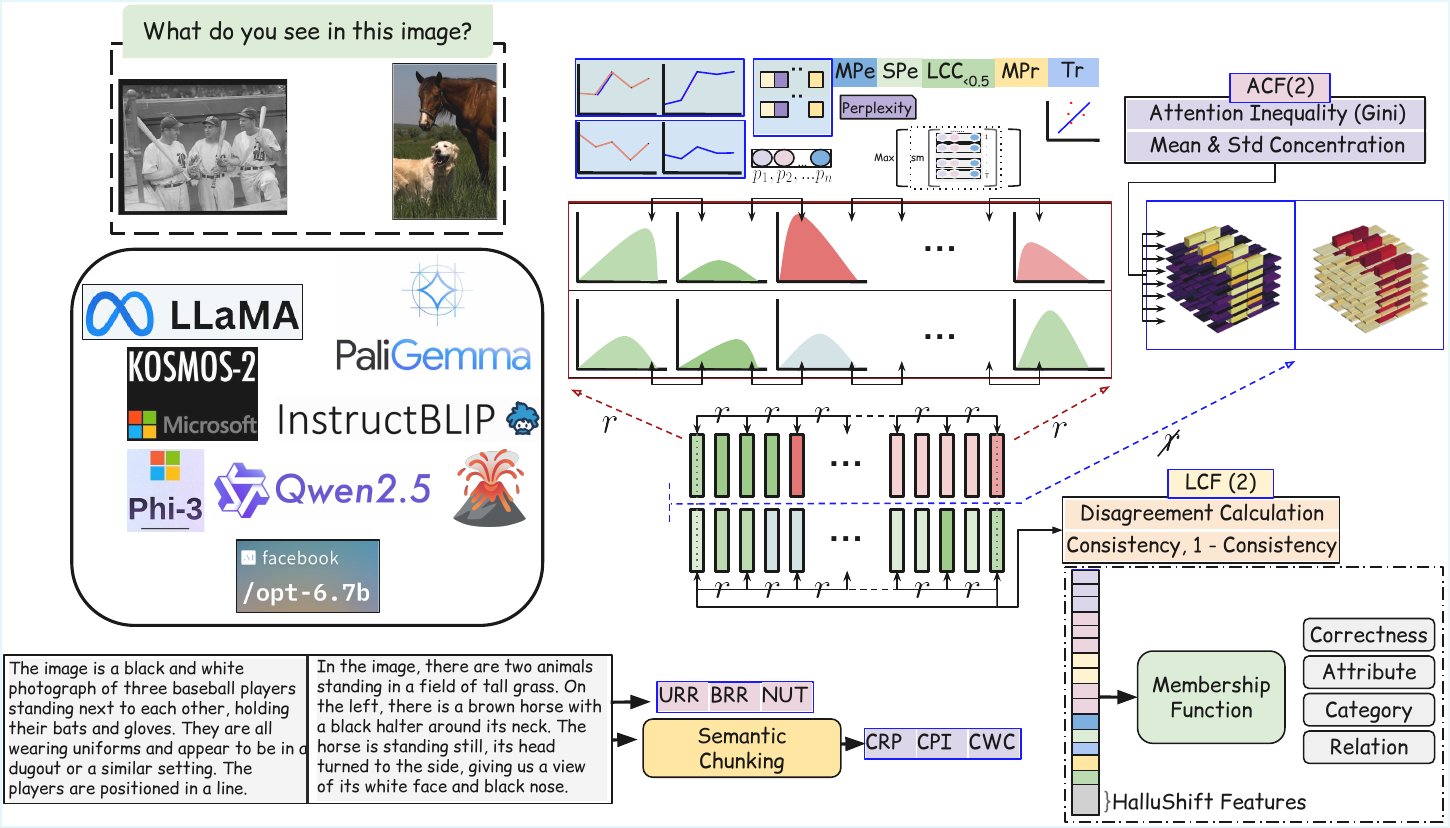}
  \caption{Illustration of our proposed \textsc{HalluShift++} framework. First, we extract features from the internal states of MLLMs including: (1) original \textsc{HalluShift} features (1-62), and (2) novel \textsc{HalluShift++} features (63-74, highlighted in blue boxes) comprising 2 attention concentration features (ACF) measuring focus dispersion via Gini coefficients, 2 layer consistency features (LCF) detecting cross-layer disagreement, and perplexity/confidence features capturing uncertainty patterns. Color transitions from green to red in internal distributions indicate hallucination-correlated patterns. Second, semantic chunking decomposes generated text into object, attribute, and relation components. Finally, our membership function processes the combined 74-dimensional feature vectors to assign hierarchical hallucination scores across four categories (Correctness, Attribute, Category, Relation) rather than binary classification.}
  \label{fig:example}
\end{figure*}
 
\section{Problem Formulation}
Let $\mathcal{M}$ be a MLLM that processes multimodal inputs to generate textual descriptions. For any input pair $(\mathbf{I}, \mathbf{x})$ where $\mathbf{I} \in \mathbb{R}^{H \times W \times C}$ represents an image and $\mathbf{x} = (x_1, \ldots, x_n), x_i \in \mathcal{V}$ is a sequence of words from vocabulary $\mathcal{V}$ constructing text prompt. The model generates a description sequence $\mathbf{y} = (y_1, \ldots, y_m) \in \mathcal{V}^m$, where m is the maximum number of output tokens and each token $y_t$ is sampled from the conditional distribution $P(y_t | \mathbf{I}, \mathbf{x}, y_{<t})$. The central challenge resides in the inherent tendency of MLLMs to generate \textit{hallucinated} content, textual descriptions that are not grounded in visual input $\mathbf{I}$. Given the stochastic nature of the generation process, where $P(\mathbf{y} | \mathbf{I}, \mathbf{x}) = \prod_{t=1}^m P(y_t | \mathbf{I}, \mathbf{x}, y_{<t})$, the model may produce semantically plausible but factually incorrect descriptions. However, the selection of $y_t$ is derived from a top-k sampling or greedy approach in the vocabulary layer that results from the internal layer distributions. Let $\mathcal{G} = \{g_1, \ldots, g_p\}$ represent the ground truth annotations that accurately describe the visual content of image $\mathbf{I}$, where $p$ is the number of ground truth annotations. We define a hallucination detection problem as learning a function $H: (\mathbf{I}, \mathbf{x}, \mathbf{y}) \rightarrow \mathcal{Y}$ that maps any generated description to a hallucination category, where $\mathcal{Y} = \{\text{Correct}, \text{Category}, \text{Attribute}, \text{Relation}\}$ represents a hierarchical taxonomy of hallucination types. 

\section{Proposed Method}
We present \textsc{HalluShift++} to address the unique challenge of visual grounding process inherent in MLLMs, where hallucination may arise from misaligned visual-linguistic representations \cite{bai2024hallucination}, by analyzing their internal states. This approach improves upon the original \textsc{HalluShift} methodology designed explicitly for LLMs, with additional feature construction, semantic decomposition that automatically parses generated descriptions into object, attribute, relation chunks and a hierarchical membership function that performs multi-class classification to detect various object hallucination, resulting in superior hallucination detection performance in both MLLMS and LLMs.


\subsection{Feature Construction}
\label{subsec:feature_extraction}
The core innovation of \textsc{HalluShift++} resides in extension towards 74-dimensional feature representation that encapsulates both established hallucination indicators \cite{dasgupta2025hallushift} and novel computational patterns. Our feature extraction function $\mathcal{F}: (\mathbf{I}, \mathbf{x}, \mathbf{y}, \mathcal{S}) \rightarrow \mathbb{R}^{74}$ operates in the internal state of the model $\mathcal{S} = \{\mathbf{h}^{(l)}, \mathbf{A}^{(l)}, \mathbf{p}^{(t)}\}_{l=1,t=1}^{L,m}$ comprising hidden states, attention weights, and token probabilities in the layers and generation steps. \\
\noindent
\textbf{Original \textsc{HalluShift} Features (1-62):} We extract 62 foundational features following \textsc{HalluShift} \cite{dasgupta2025hallushift}. It combines distributional shift analysis of internal layers with token probability based detection. \\
\noindent
\textbf{Enhanced \textsc{HalluShift++} Features (63-74):} While the original \textsc{HalluShift} features effectively capture distributional shifts in text-only LLMs, the MLLMs present unique challenges that arises from the need to bridge visual and linguistic modalities. Unlike LLMs that operate within a single textual domain with consistent semantic spaces, MLLMs must \textit{ground} text generation in visual evidence, while managing the inherent \textit{modality gap} between continuous visual representations and linguistic tokens. This introduces three unique failure modes absent in text-only scenarios: 1) \textit{Cross-layer inconsistency} occurs when visual features extracted by early vision encoders become progressively distorted as they pass through language decoder layers, causing the model to loose visual grounding and generate output inconsistent with the image, 2) \textit{Attention dispersion} occurs when models trained on datasets with noisy image-text alignments learn to rely heavily on language priors rather than visual evidence, causing cross-modal attention to become unfocused across irrelevant visual regions, especially in domain adaptation scenarios.


3) \textit{Confidence degradation} emerges from uncertainty propagation across the vision-language integration pipeline, where ambiguous visual interpretations cascade through text generation, resulting in overconfident but incorrect descriptions. To address these MLLM specific failure modes, we introduce 12 novel features: \\
\noindent
\textit{Layer Consistency Features, LCF (63-64):} Detect \textit{Cross-layer inconsistency} by measuring the alignment between early and late text decoder layers. When model hallucinate, there tends to be greater disagreement between what early layers represent vs what late layers represent for same visual-linguistic input.
\begin{align}
c = \left[\text{cosine\_sim}(\mathbf{h}^{(early)}, \mathbf{h}^{(late)}) + 1\right] / 2, \quad \mathcal{F}_{\text{consistency}} = [c, 1-c]
\end{align}
where $\mathbf{h}^{(early)}$ and $\mathbf{h}^{(late)}$ are flattened hidden states from specific text decoder layers. The early layer index is $\min(\text{text\_start} + 5, \text{max\_layer} - 10)$ (5th text decoder layer or 10 layers before the end), and the late layer index is $\text{max\_layer} - 2$ (2nd to last layer). We normalize the cosine similarity from [-1,1] to [0,1] range and extract both consistency ($c$) and inconsistency ($1-c$) as complementary features to provide the classifier with dual perspectives on layer alignment for optimal decision boundaries. \\

\noindent
\textit{Attention Concentration Features, ACF (64-65):} We address \textit{Attention dispersion} by measuring attention inequality using Gini coefficients, higher values indicate desirable concentrated attention on specific visual regions, while lower values signal problematic scattered attention associated with hallucination. For each attention layer, we compute the Gini coefficient as:
\begin{equation}
G_l = \left[2\sum_{i=1}^{n}\sum_{j=1}^{i} s_{j}\right] / \left[n\sum_{i=1}^{n} s_{i}\right] - 1,
\end{equation}
\begin{equation}
    \quad \mathcal{F}_{\text{concentration}} = \left[\text{mean}(\{|G_l|\}_{l=-3}^{-1}), \text{std}(\{|G_l|\}_{l=-3}^{-1})\right]
\end{equation}
where $s_i$ and $s_j$ both refer to elements of the same sorted flattened attention weights for layer $l$, $n$ is the total number of attention weights, and $G_l$ is the Gini coefficient measuring concentration. When attention is properly focused (grounded generation), most weights concentrate on few relevant regions yielding high Gini values ($G_l \rightarrow 1$). When attention disperses across many irrelevant regions (hallucination-prone), weights distribute more uniformly yielding low Gini values ($G_l \rightarrow 0$). We analyze the last 3 attention layers and compute mean concentration level and consistency deviation (std) to detect this hallucination-predictive attention dispersion pattern. \\

\noindent
\textit{Perplexity and Confidence Features (67-71):} Address \textit{Confidence degradation} by detecting when models generate text with deteriorating certainty, indicating potential hallucination due to weak visual grounding. For each generated token position $t \in \{1, 2, \ldots, T\}$, we extract logits $\mathbf{l}_t$ containing raw scores for the entire vocabulary. We convert this to probabilities and extract the maximum probability.
\begin{align}
\mathbf{p}_t = \text{softmax}(\mathbf{l}_t[0]), \quad p_{\max}^{(t)} = \max(\mathbf{p}_t)
\end{align}
where, $\mathbf{l}_t[0]$ selects the first batch dimension, and $p_{\max}^{(t)}$ represents the model's confidence in the chosen token.

\noindent
Perplexity is computed as the reciprocal of maximum probability,
\begin{align}
\text{ppl}_t = 1 / {p_{\max}^{(t)}}
\end{align}

\noindent
After processing all $T$ token positions, we construct two lists, $\{p_{\max}^{(1)}, \\ \ldots, p_{\max}^{(T)}\}$ (maximum probabilities) and $\{\text{ppl}_1, \ldots, \text{ppl}_T\}$ (perplexities). From these, we extract 5 features:

\noindent
\textbf{Mean Perplexity (MPe):} Captures the average model uncertainty across all generated tokens. Higher values indicate lower overall confidence in the generated content.
\begin{align}
f_1 = \frac{1}{T} \sum_{t=1}^{T} \text{ppl}_t
\end{align}

\noindent
\textbf{Perplexity Standard Deviation (SPe):} Measures the consistency of model confidence throughout generation. High variance suggests unstable confidence patterns associated with hallucination.
\begin{align}
f_2 = \sqrt{\frac{1}{T-1} \sum_{t=1}^{T} (\text{ppl}_t - f_1)^2}
\end{align}

\noindent
\textbf{Confidence Trend (Tr):} Uses linear regression to detect whether model confidence increases or decreases over the sequence. Negative slopes indicate deteriorating confidence, signaling potential hallucination.
\begin{align}
f_3 = \frac{T\sum_{t=1}^{T}(t-1) \cdot p_{\max}^{(t)} - \sum_{t=1}^{T}(t-1) \sum_{t=1}^{T}p_{\max}^{(t)}}{T\sum_{t=1}^{T}(t-1)^2 - \left(\sum_{t=1}^{T}(t-1)\right)^2}
\end{align}

\noindent
\textbf{Mean Confidence (MPr):} Represents the average maximum probability across all token positions, providing a direct measure of model certainty.
\begin{align}
f_4 = \frac{1}{T} \sum_{t=1}^{T} p_{\max}^{(t)}
\end{align}

\noindent
\textbf{Low-Confidence Token Fraction (LCC):} Counts the proportion of tokens with maximum probability $p_{\max}^{(t)}$ below 0.5, indicating systematic uncertainty in generation.
\begin{align}
f_5 = \frac{|\{t : p_{\max}^{(t)} < 0.5\}|}{T}
\end{align}

These five features collectively capture different aspects of confidence degradation that correlate with hallucination in MLLMs, where weak visual grounding leads to uncertain text generation patterns.

\noindent
\textit{Token Pattern Features (72-74):} 3 token pattern features are constructed to detect when model generate text that lacks visual grounding by analyzing repetition and diversity patterns that differ from naturally grounded descriptions.
\begin{align}
URR = 1 - |U| / |T|, \quad BRR = 1 - |B_u| / |B|, \quad NUT = |U| / |T|
\end{align}
where, $URR$ represents unique repetition ratio, $BRR$ bi-gram repetition ratio, $NUT$ normalized unique tokens, $U$ is the set of unique tokens, $T$ is all tokens, $B_u$ is unique bigrams, and $B$ is all bigrams in the generated text. Visually grounded descriptions exhibit natural diversity and varied phrasing, while hallucinated content shows repetitive patterns as models use memorized language patterns. High repetition ratios ($URR, BRR \rightarrow 1$) and low diversity ($NUT \rightarrow 0$) indicate over-reliance on language priors rather than visual evidence.

\begin{algorithm}
\caption{Semantic Chunk Extraction}
\KwIn{Text string $T$}

$D \leftarrow \texttt{NLP}(T)$; \quad $\mathcal{C} \leftarrow \emptyset$\;

\ForEach{$t \in D$}{
  \If{$t.\texttt{pos} \in \{\texttt{NOUN}, \texttt{PROPN}\}$ \textbf{and} $t$ not stopword}{
    Add object chunk for $t$ to $\mathcal{C}$\;
  }
  \If{$t.\texttt{pos} = \texttt{ADJ}$ \textbf{and} $t.\texttt{head.pos} \in \{\texttt{NOUN}, \texttt{PROPN}\}$}{
    Add attribute chunk for $(t, t.\texttt{head})$ to $\mathcal{C}$\;
  }
  \If{$t.\texttt{dep} \in \{\texttt{prep}, \texttt{agent}\}$ \textbf{or} $t.\texttt{pos} = \texttt{VERB}$}{
    $\mathcal{O} \leftarrow$ nouns among $t$'s children\;
    \If{$|\mathcal{O}| \geq 2$}{
      Add relation chunk for $(\mathcal{O}[0], t, \mathcal{O}[1])$ to $\mathcal{C}$\;
    }
  }
}

Remove duplicate/short chunks from $\mathcal{C}$; sort by start index\;

\Return{$\mathcal{C}$}
\end{algorithm}
\subsection{Semantic Chunking Strategy}
\label{subsec:semantic_chunking}
A fundamental limitation of the original \textsc{HalluShift} approach lies in its treatment of generated text as monolithic units, assigning binary hallucination labels to entire responses, which fails at places where hallucinations manifest at different semantic granularities within the same response. This necessitates a semantic decomposition that can isolate and evaluate individual units independently. To address this, we implement a semantic decomposition approach, adapting \cite{park2025halloc} for Vision-Language tasks. Specifically, we extract three types of chunks from generated text: \textit{object mentions} (individual nouns and proper nouns representing entities in the description), \textit{attribute-object pairs} (adjective-noun combinations capturing descriptive properties) and relationship patterns (preposition/verb based connections between multiple objects that encode spatial or action-based relations). Each extracted chunk is then characterized by three core positional and structural features: Chunk Word Count (CWC: number of tokens in the chunk), Chunks Per Image (CPI: the total chunk count $k$ for that image), and Chunk Relative Position (CRP: computed as $i/k$, where $i$ is the position of the current chunk) capturing semantic density, descriptive granularity, and temporal ordering respectively.

\subsection{Hierarchical Ground Truth Matching}
\label{subsec:ground_truth_matching}
The critical challenge for training data preparation becomes determining whether each chunk represents factual content grounded in the image or hallucinated content to generate supervised learning labels. Traditional approaches such as CHAIR \cite{bordes2024chair} rely on simple keyword matching against object vocabularies, that fails to capture the nuanced relationships between objects, their properties, and spatial arrangements. Our hierarchical ground truth matching system addresses this limitation by implementing a verification process that respects the inherent hierarchy, if an object doesn't exist in the image (category error), then discussing its attributes or relationships is meaningless. Conversely, if an object exists but has wrong attributes, the spatial relationships might still be partially correct. This motivates our three-stage verification process.
\noindent
\textbf{Ground Truth Extraction Functions:} We define three extraction functions that parse ground truth annotations $\mathcal{G}$ to create searchable representations. Firstly, the object extraction function, processes caption to identify all mentioned entities using named entity recognition \cite{keraghel2024ner} and noun phrase extraction, returning a set of canonical object labels $\{\text{person}, \text{car}, ...\}$. Secondly, the attribute extraction function considers object labels to analyze dependency relationships for object-attribute pairs, creating tuples $\{(\text{car}, \text{red}), (\text{building}, \text{tall}), ...\}$ where each attribute is linked to its corresponding object. Thirdly, the relation extraction function identifies spatial and action relationships through prepositional phrase analysis and verb-argument structures, generating relational triplets $\{(\text{car}, \text{parked-next-to}, \text{building}), ...\}$. To delineate this, we utilize predefined lexical dictionaries including sequence of objects, attribute categories (i.e., color, size, shape, condition, material) and spatial-action relations.\\
\noindent
\textbf{Chunk Analysis Functions:} After processing ground truths, we extract list of generated chunks $\mathbf{c}$ from generated text $\mathbf{y}$. To expedite ground truth matching, we define three chunk analysis functions that extract comparable information from generated chunks. Here, the object function identifies the primary entity within each generated chunk $\mathbf{c}_i$ using the same NLP techniques applied to ground truth. Similarly, the attribute function forms attribute claims from the chunk, while relation function identifies relational assertions made within the chunk.\\
\noindent
This hierarchical approach ensures that category-level errors (non-existent objects) take precedence over attribute-level errors (incorrect properties), which in turn take precedence over relation-level errors (incorrect spatial/action relationships).
\subsection{Membership Function}
\label{subsec:Membership Function}
Following \textsc{HalluShift} \cite{dasgupta2025hallushift} membership function approach, we extend the framework to handle multiple hallucination types. In contrast to the original \textsc{HalluShift} that performs binary classification (hallucinated vs. truthful), \textsc{HalluShift++} implements a membership function that assigns probabilistic scores across four classes: $\mathcal{Y} = \{\text{Correct}, \text{Category}, \text{Attribute}, \text{Relation}\}$. We combine the 74-dimensional, \textsc{HalluShift++} features with three additional chunk-specific contextual features $\phi(c_i) \in \mathbb{R}^{3}$ (CWC, CRP, CPI), where $\mu(c_i, y_j) = f_\theta([\mathcal{F}_{\text{\textsc{HalluShift++}}}(c_i) \in \mathbb{R}^{74}; \phi(c_i) \in \mathbb{R}^{3}])_j$ employs attention mechanisms on these 3 chunk features to weigh feature importance dynamically across different hallucination types.

\section{Experimental Analysis}
\subsection{Dataset Overview}
\label{subsec:dataset_setup}
We evaluate \textsc{HalluShift++} on two prominent vision-language benchmarks: MS-COCO \cite{lin2015microsoft} and LLaVA \cite{liu2023visualinstructiontuning}. MS-COCO provides a diverse collection of natural images with detailed captions, enabling comprehensive evaluation of object recognition, attribute assignment, and spatial relationship understanding. Contrastingly, LLaVA benchmark focuses on instruction-following capabilities in vision-language tasks, providing a challenging testbed for hallucination detection in conversational contexts.

\subsection{Experimental Setup}
\label{subsec:training_config}
We engage eight state-of-the-art MLLMs with several architectural paradigms: Llama-3.2-11B-Vision-Instruct and Llama-3.2-11B-Vision \cite{touvron2023llama}, PaliGemma-3B-Mix-224 \cite{beyer2024paligemma}, Kosmos-2-Patch14-224 \cite{pan2024kosmos}, InstructBLIP-Vicuna-7B \cite{dai2023instructblip}, Phi-3.5-Vision-Instruct \cite{abdin2024phi3}, Qwen 2.5-VL-3B-Instruct \cite{bai2023qwentechnicalreport}, and LLaVA-1.5-7B-HF \cite{liu2023visualinstructiontuning}. 
For consistent evaluation, we generate model responses using greedy decoding with a maximum output length of 64 tokens. Ground truth annotations are established through careful manual verification against dataset annotations, with hallucination labels assigned according to our hierarchical taxonomy: Category (non-existent objects), Attribute (incorrect properties), Relation (spatial/action inconsistencies), and Correct (factually accurate). We extract both \textsc{HalluShift} and \textsc{HalluShift++} features from internal layers of the model with range $2$. We employ a three-layer neural architecture with progressive feature integration as the membership estimator. 
We use AdamW optimizer with an initial learning rate of 1e-4 and cosine annealing schedule with batch size 32. \\
\noindent
Training employs early stopping based on validation AUC-ROC with patience of 10 epochs. We apply class balancing techniques like SMOTE \cite{chawla2002smote} and class weighting to handle potential dataset imbalances. SMOTE generates synthetic samples for minority classes, while enhanced class weighting applies higher penalties to misclassifications of rare hallucination types, particularly addressing the severe imbalance in attribute hallucinations which constitute less than 5\% of samples. All experiments are conducted on NVIDIA A40 GPU with 48 GB VRAM using mixed-precision training for computational efficiency.
\noindent
\begin{algorithm}
\caption{Ground Truth-Aware Chunk Classification}
\KwIn{Chunk $c$, Ground Truth Set $\mathcal{G}$}
\KwOut{Label $\mathcal{L} \in \{$\texttt{CORRECT}, \texttt{CATEGORY\_HALLUC}, \texttt{ATTRIBUTE\_HALLUC}, \texttt{RELATION\_HALLUC}$\}$}

Extract $\mathcal{O} \leftarrow$ objects from $\mathcal{G}$\;
Extract $\mathcal{A} \leftarrow$ attributes from $\mathcal{G}$\;
Extract $\mathcal{R} \leftarrow$ relations from $\mathcal{G}$\;

\eIf{$c.\texttt{type} = \texttt{object}$}{
  \If{$c.\texttt{text} \notin \mathcal{O}$ (match)}{
    \Return{\texttt{CATEGORY\_HALLUC}}
  }
}{
  Let $objs \leftarrow$ associated objects in $c$ (1 or 2)\;
  \If{any $o \in objs$ not in $\mathcal{O}$}{
    \Return{\texttt{CATEGORY\_HALLUC}}
  }
  \eIf{$c.\texttt{type} = \texttt{attribute}$}{
    \If{$c.\texttt{attribute} \notin \mathcal{A}$ (match/sim)}{
      \Return{\texttt{ATTRIBUTE\_HALLUC}}
    }
  }{
    \If{$c.\texttt{relation} \notin \mathcal{R}$ (match/sim)}{
      \Return{\texttt{RELATION\_HALLUC}}
    }
  }
}

\Return{\texttt{CORRECT}}
\end{algorithm}

\section{Comparative Analysis}
\subsection{Performance of Vision-Language Tasks}
Table~\ref{tab:main_results} presents a comprehensive comparison of \textsc{HalluShift++} against the original \textsc{HalluShift} method across multiple MLLMs and datasets. We didn't compare with external evaluation metrics such as POPE \cite{li2023evaluating} and FAITHScore \cite{jing2023faithscore} because our study focuses exclusively on methods that leverage internal model representations. Our results reveal several critical insights about hallucination detection in MLLMs. \textsc{HalluShift++} achieves consistent improvements of 27.7 - 64.1\% AUC-ROC gains across all tested models, indicating that our multimodal specific features capture fundamental hallucination patterns independent of architectural design. This consistency validates our hypothesis that cross-layer inconsistency, attention dispersion, and confidence degradation represent universal failure modes in multimodal systems.
\noindent
\begin{table*}[!ht]
\centering
\caption{Comparative analysis of hallucination detection methods. Metrics reported include AUC-ROC, precision, and f1-score.}
\label{tab:main_results}
\scriptsize
\resizebox{\textwidth}{!}{%
\begin{tabular}{@{}llccccc@{}}
\toprule
\textbf{Model} & \textbf{Dataset} & \textbf{Method} & \textbf{AUC\_ROC (\%)} & \textbf{Precision (\%)} & \textbf{F1-Score (\%)} \\
\midrule
\multirow{6}{*}{\textbf{Llama-3.2-11B-Vision-Instruct}} & \multirow{3}{*}{mscoco} & \textsc{ITI} & 49.80 & 48.30 & 57.20 \\
 & & \textsc{HalluShift} & 52.10 & 52.10 & 60.50 \\
 & & \textsc{HalluShift++} & \textbf{86.10} & \textbf{77.40} & \textbf{62.60} \\
\cline{2-6}
 & \multirow{3}{*}{llava} & \textsc{ITI} & 48.20 & 31.50 & 40.10 \\
 & & \textsc{HalluShift} & 51.60 & 34.30 & 43.30 \\
 & & \textsc{HalluShift++} & \textbf{82.80} & \textbf{67.90} & \textbf{59.60} \\
\midrule
\multirow{6}{*}{\textbf{paligemma-3b-mix-224}} & \multirow{3}{*}{mscoco} & \textsc{ITI} & 58.40 & 62.10 & 61.50 \\
 & & \textsc{HalluShift} & 61.70 & 65.40 & 64.20 \\
 & & \textsc{HalluShift++} & \textbf{87.00} & \textbf{77.50} & \textbf{69.00} \\
\cline{2-6}
 & \multirow{3}{*}{llava} & \textsc{ITI} & 57.90 & 54.30 & 50.20 \\
 & & \textsc{HalluShift} & 61.70 & 57.20 & 53.00 \\
 & & \textsc{HalluShift++} & \textbf{83.80} & \textbf{70.60} & \textbf{61.80} \\
\midrule
\multirow{6}{*}{\textbf{kosmos-2-patch14-224}} & \multirow{3}{*}{mscoco} & \textsc{ITI} & 50.30 & 40.80 & 38.20 \\
 & & \textsc{HalluShift} & 54.00 & 43.70 & 41.00 \\
 & & \textsc{HalluShift++} & \textbf{85.40} & \textbf{67.20} & \textbf{57.60} \\
\cline{2-6}
 & \multirow{3}{*}{llava} & \textsc{ITI} & 51.80 & 40.20 & 39.50 \\
 & & \textsc{HalluShift} & 55.40 & 43.40 & 42.30 \\
 & & \textsc{HalluShift++} & \textbf{85.90} & \textbf{68.30} & \textbf{63.40} \\
\midrule
\multirow{4}{*}{\textbf{instructblip-vicuna-7b}} & \multirow{2}{*}{mscoco} & \textsc{HalluShift} & 51.70 & 34.30 & 43.30 \\
 & & \textsc{HalluShift++} & \textbf{85.00} & \textbf{67.70} & \textbf{56.40} \\
\cline{2-6}
 & \multirow{2}{*}{llava} & \textsc{HalluShift} & 51.50 & 25.00 & 33.30 \\
 & & \textsc{HalluShift++} & \textbf{84.80} & \textbf{67.60} & \textbf{61.50} \\
\midrule
\multirow{4}{*}{\textbf{Phi-3.5-vision-instruct}} & \multirow{2}{*}{mscoco} & \textsc{HalluShift} & 54.90 & 45.90 & 54.30 \\
 & & \textsc{HalluShift++} & \textbf{85.30} & \textbf{71.40} & \textbf{56.90} \\
\cline{2-6}
 & \multirow{2}{*}{llava} & \textsc{HalluShift} & 58.40 & 47.50 & 49.50 \\
 & & \textsc{HalluShift++} & \textbf{84.30} & \textbf{64.80} & \textbf{58.30} \\
\midrule
\multirow{4}{*}{\textbf{Qwen2.5-VL-3B-Instruct}} & \multirow{2}{*}{mscoco} & \textsc{HalluShift} & 55.90 & 61.10 & 56.20 \\
 & & \textsc{HalluShift++} & \textbf{83.60} & \textbf{72.90} & \textbf{58.50} \\
\cline{2-6}
 & \multirow{2}{*}{llava} & \textsc{HalluShift} & 55.50 & 51.70 & 44.80 \\
 & & \textsc{HalluShift++} & \textbf{84.00} & \textbf{66.80} & \textbf{59.60} \\
\midrule
\multirow{4}{*}{\textbf{Llama-3.2-11B-Vision}} & \multirow{2}{*}{mscoco} & \textsc{HalluShift} & 65.80 & 52.50 & 60.90 \\
 & & \textsc{HalluShift++} & \textbf{91.70} & \textbf{83.00} & \textbf{71.80} \\
\cline{2-6}
 & \multirow{2}{*}{llava} & \textsc{HalluShift} & 71.10 & 53.20 & 61.60 \\
 & & \textsc{HalluShift++} & \textbf{88.20} & \textbf{76.40} & 52.20 \\
\midrule
\multirow{4}{*}{\textbf{llava-1.5-7b-hf}} & \multirow{2}{*}{mscoco} & \textsc{HalluShift} & 53.10 & 42.10 & 43.10 \\
 & & \textsc{HalluShift++} & \textbf{84.70} & \textbf{67.20} & \textbf{58.40} \\
\cline{2-6}
 & \multirow{2}{*}{llava} & \textsc{HalluShift} & 58.80 & 74.00 & 79.50 \\
 & & \textsc{HalluShift++} & \textbf{96.50} & \textbf{94.80} & \textbf{88.60} \\
\bottomrule
\end{tabular}%
}
\end{table*}
\noindent
\noindent
A particularly compelling aspect of our results is the scale agnostic effectiveness of \textsc{HalluShift++}. Even a very small model like Kosmos-2-Patch14-224 \cite{pan2024kosmos} with only 1.66B params show $\sim58\%$ performance gain in AUC-ROC, while models with 7B params like LLaVA-1.5-7B-HF \cite{liu2023visualinstructiontuning} and InstructBLIP-Vicuna-7B \cite{dai2023instructblip} have shown a significant improvement of $\sim64\%$ than base \textsc{HalluShift} counterpart. These smaller models demonstrate that effective hallucination detection does not require massive computation resources, making our approach practical for resource constrained environments. Simultaneously, larger models with 11B params like Llama-3.2-11B-Vision \cite{touvron2023llama} achieve 91.7\% AUC-ROC, showcasing the effectiveness of \textsc{HalluShift++} across the model parameter spectrum.

\begin{table}[!ht]
\centering
\caption{Performance comparison between \textsc{HalluShift} \cite{dasgupta2025hallushift}, \textsc{HalluShift++}, HaloScope \cite{du2024haloscope}, and CCS* \cite{burns2022discovering} on LLMs for question-answering tasks.}
\label{tab:llm_qa_results}
\resizebox{0.4\textwidth}{!}{%
\begin{tabular}{@{}llcc@{}}
\toprule
\textbf{Model} & \textbf{Dataset} & \textbf{Method} & \textbf{AUC\_ROC} \\
\midrule
\multirow{8}{*}{\textbf{OPT-6.7B}} & \multirow{4}{*}{TruthfulQA} & CCS* & 63.91 \\
 &  & HaloScope & 73.17 \\
 & &  \textsc{HalluShift} & \underline{89.91} \\
 &  & \textsc{HalluShift++} & \textbf{92.68} \\
\cline{2-4}
 & \multirow{4}{*}{TyDiQA-GP} & CCS* & 64.62\\
 &  & HaloScope & 80.98 \\
 &  & \textsc{HalluShift} & \underline{85.11} \\
 &  & \textsc{HalluShift++} & \textbf{87.58} \\
\midrule
\multirow{8}{*}{\textbf{LLaMA-2-7B}} & \multirow{4}{*}{TruthfulQA} & CCS* & 67.95 \\
 &  & HaloScope & 78.64 \\
 &  & \textsc{HalluShift} & \underline{89.93} \\
 &  & \textsc{HalluShift++} & \textbf{90.82} \\
\cline{2-4}
 & \multirow{4}{*}{TyDiQA-GP} & CCS* & 80.38 \\
 &  & HaloScope & \textbf{94.04} \\
 &  & \textsc{HalluShift} & 87.61 \\
 &  & \textsc{HalluShift++} & \underline{89.66} \\
\bottomrule
\end{tabular}%
}
\end{table}

\subsection{Performance on Language Model QA Tasks}
\label{subsec:llm_qa_performance}
To demonstrate the generalizability of \textsc{HalluShift++} beyond vision-language tasks, we evaluate our approach on text-only question-answering scenarios using large language models. Table~\ref{tab:llm_qa_results} presents comparative results on TruthfulQA and TyDiQA datasets using OPT-6.7B and LLaMA-2-7B models. \\
\noindent
The results demonstrate that \textsc{HalluShift++} maintains its effectiveness in text-only settings, achieving consistent improvements over the baseline \textsc{HalluShift} method. The most substantial improvement is observed for OPT-6.7B on TruthfulQA (2.77 point improvement), while more modest but consistent gains are achieved across other configurations. These results validate that our enhanced feature extraction methodology generalizes effectively beyond multimodal scenarios, suggesting broader applicability for hallucination detection in diverse language model architectures and task domains.

\section{Ablation Study}
\subsection{Feature Importance Analysis}
\label{subsec:detailed_feature_importance}
\begin{figure}[!ht]
\centering
\includegraphics[width=0.95\columnwidth]{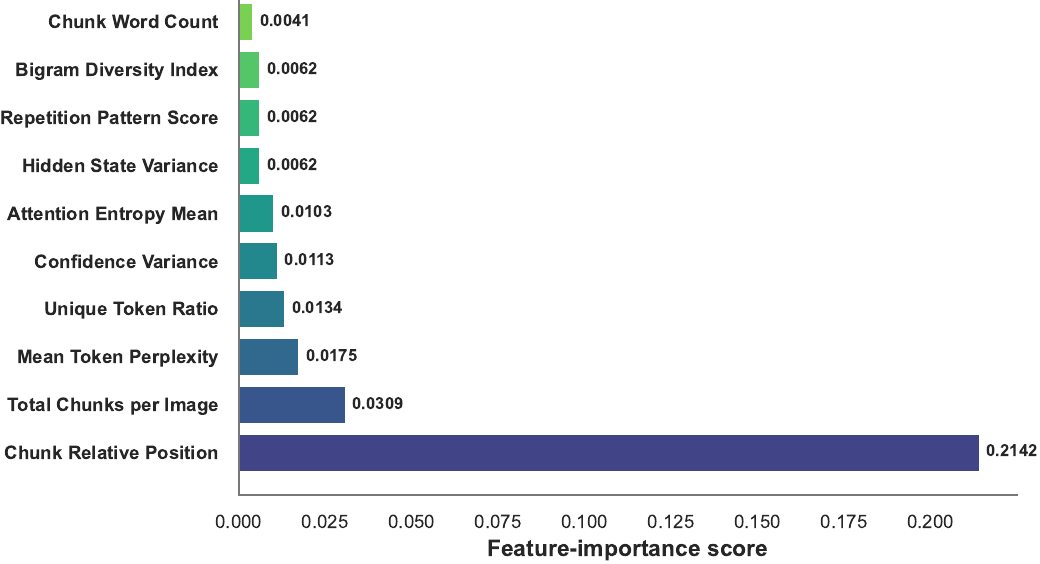}
\caption{Feature importance analysis showing top 10 features for the best performing model (Llama-3.2-11B-Vision). The horizontal bar chart shows the contribution of each feature to hallucination detection accuracy, with semantic context features dominating the rankings while \textsc{HalluShift++} features provide significant improvements. Feature names are interpretable rather than numerical for clarity.}
\label{fig:feature_importance}
\end{figure}

We utilize best performing model, Llama-3.2-11B-Vision, for further analysis of feature importance, providing interpretable feature names and their contributions to hallucination detection (Fig~\ref{fig:feature_importance})
\noindent
Our detailed analysis reveals that \textit{Chunk Relative Position} emerges as the most essential feature with an importance score of $+0.2142$, indicating that the location of content within generated responses is highly predictive of hallucination. This finding suggests that hallucinated content tends to appear in specific positions within model outputs, potentially reflecting patterns in the generation process. The second most dominant feature, \textit{Total Chunks per Image} ($+0.0309$), captures the complexity of generated descriptions. 
Models generating overly detailed descriptions may be more prone to hallucination, as they attempt to provide information beyond what can be reliably extracted from the visual input. \textsc{HalluShift++} effectively captures this phenomenon through the attention concentration features detecting when model lose focus across too many visual regions while attempting detailed descriptions. Additionally, the confidence degradation features identify increasing uncertainty as models venture beyond reliable visual evidence. \textit{Mean Token Perplexity} ($+0.0175$) shows strong contribution cause models exhibit higher uncertainty when generating content not grounded in visual evidence compared to describing what they actually observe in images. Fruthermore, the \textit{Unique Token Ratio} ($+0.0134$) proves effective as models exhibit repetitive patterns based memorized language prior from training data, when lacking visual grounding. 

\subsection{Semantic Chunking Effectiveness}
\label{subsec:chunking_effectiveness}
We further evaluate the impact of our semantic chunking strategy on detection performance, demonstrating the value of semantic decomposition in Table~\ref{tab:chunking_effectiveness}.

\begin{table}[!ht]
\centering
\caption{Semantic chunking strategy effectiveness showing performance gains from hierarchical text decomposition.}
\label{tab:chunking_effectiveness}
\resizebox{0.4\textwidth}{!}{%
\begin{tabular}{@{}lccc@{}}
\toprule
\textbf{Chunking Strategy} & \textbf{Avg. Chunks/Image} & \textbf{AUC-ROC} \\
\midrule
No Chunking & 1.0 & 0.612 \\
Sentence-level & 2.3 & 0.698 \\
Object-only & 8.7 & 0.743 \\
Object + Attribute & 12.1 & 0.798 \\
Complete Semantic & 14.2 & 0.847 \\
\bottomrule
\end{tabular}%
}
\end{table}

\section{Conclusion}
In this work, we introduced \textsc{HalluShift++}, a framework that advances hallucination detection from traditional binary classification to hierarchical understanding of multimodal hallucination patterns. By extending the distributional shift analysis from text-only LLMs to the complex visual-linguistic landscape of MLLMs, we demonstrated that hallucinations manifest as quantifiable perturbations across multiple dimensions: cross-layer inconsistency, attention dispersion, and confidence degradation, phenomena that remain invisible to conventional approaches. 


\end{document}